\documentclass[12pt]{article}

\usepackage{booktabs}
\usepackage{siunitx}
\newcommand{\RyanEdits}[1]{{\color{red}#1}}

\newcommand{\hr}{\mathbf{x^{HR}}}
\newcommand{\hathr}{\mathbf{\hat{x}^{HR}}}
\newcommand{\lr}{\mathbf{x^{LR}}}

\newcommand{\rs}{\mathbf{r}}
\newcommand{\bi}{\mathbf{I}}
\newcommand{\bz}{\mathbf{0}}
\usepackage{etoolbox}
\usepackage{bm}

\usepackage{arydshln}

\makeatletter
\def\mathcolor#1#{\@mathcolor{#1}}
\def\@mathcolor#1#2#3{%
	\protect\leavevmode
	\begingroup
	\color#1{#2}#3%
	\endgroup
}
\makeatother

\usepackage{url}
\usepackage{setspace}
\usepackage{algorithm}
\usepackage{algpseudocode}
\algrenewcommand\algorithmicrequire{\textbf{Input:}}
\algrenewcommand\algorithmicensure{\textbf{Output:}}

\usepackage[table,xcdraw]{xcolor}

\usepackage{graphicx}
\usepackage{multirow}

\graphicspath{ {./figures/} }

\usepackage{caption}
\usepackage{subcaption}

\usepackage{amsmath} 
\usepackage{amsfonts}
\usepackage{nicefrac}

\usepackage[a4paper, total={6in, 8in}]{geometry}

\usepackage[symbol]{footmisc}

\usepackage{multirow}

\usepackage[
colorlinks=true,
linkcolor=blue,
filecolor=blue,
urlcolor=blue,
citecolor=blue,
pagebackref=true
]{hyperref}
\makeatletter
\renewcommand*{\backref}[1]{}
\renewcommand*{\backrefalt}[4]{%
	\ifcase #1
	(Not cited.)%
	\or
	\textsuperscript{#2}%
	\else
	\textsuperscript{#2}%
	\fi}
\makeatother

\begin{document}
	\begin{center}	
\textbf{\Large {MRI super-resolution in ten sampling steps using a diffusion bridge model}} \\
Mojtaba Safari$ ^1 $, Hang Yu$ ^2 $, Zach Eidex$ ^3 $, Mingzhe Hu$ ^3 $, Ryan J. Sanford$ ^{3,4} $, Alexandru Florea$ ^{3,5} $, Shansong Wang$ ^1 $, Chih-Wei Chang$ ^3 $, Erik H Middlebrooks$ ^6 $,  Aditya Juloori$ ^1 $, Stanley L. Liauw$ ^1 $, Ralph Weichselbaum$ ^1 $, and Xiaofeng Yang$ ^{1,3,5, \ddagger} $ \\
    \end{center}
\begin{flushleft}
$ ^1 $Department of Radiation and Cellular Oncology, The University of Chicago, Chicago, IL, United States,\\	
$ ^2 $Department of Physics and Astronomy, University of North Carolina at Chapel Hill, Chapel Hill, NC, United States,\\
$ ^3 $Department of Radiation Oncology and Winship Cancer Institute, Emory University, Atlanta, GA, United States,\\
$ ^4 $Medical Physics Graduate Program, Georgia Institute of Technology, Atlanta, GA, United States,\\
$ ^5 $Wallace H. Coulter Department of Biomedical Engineering, Georgia Institute of Technology and Emory University, Atlanta, GA, United States,\\
$ ^6 $Department of Radiology, Mayo Clinic, Jacksonville, FL, United States of America.\\

\vspace{1 cm}

$ ^\ddagger  $Corresponding Author: email: \url{xfyang@uchicago.edu}
\end{flushleft}

\newpage

\begin{abstract}
	\noindent\textbf{\textit{Objective.}} MRI provides excellent soft-tissue contrast, but long acquisition times can cause patient discomfort and lead to motion artifacts, forcing a trade-off between spatial resolution and scan time. Diffusion-based super-resolution (SR) reconstructs high-resolution (HR) images from low-resolution (LR) inputs, but typically needs many sampling steps and initializes from a Gaussian prior ill-suited to image restoration. We developed an efficient diffusion framework that reconstructs HR MRI directly from LR data.
    
    \noindent\textbf{\textit{Approach.}} We propose super-resolution diffusion bridge model (SR-DBM), a super-resolution diffusion bridge model that casts SR as a stochastic transport between the LR and HR image distributions.  Through a Doob's $h$-transform of a mean-reverting stochastic differential equation, SR-DBM pins the process to the paired HR and LR images at its endpoints, initializing reconstruction from the measured anatomy rather than from Gaussian noise. The HR image is recovered by a deterministic reverse trajectory in which a network predicts the clean image at each of only ten sampling steps. We evaluated SR-DBM on ultra-high-field 7T brain T1 MP2RAGE maps and pelvic T2-weighted prostate images against nine comparison methods using PSNR, SSIM, GMSD, and LPIPS.
    
    \noindent\textbf{\textit{Main results.}} SR-DBM attained the highest PSNR and SSIM and the lowest GMSD on both datasets (brain: $27.66\pm1.52$ dB, $0.96\pm0.02$, $7.96{\pm 1.86}$; prostate: $27.87\pm2.29$ dB, $0.80\pm0.05$, $8.38{\pm 1.44}$), with statistically significant gains over every comparison method (two-sided Wilcoxon signed-rank test with Holm correction, $p<0.05$). The strongest baseline, SR-EMamba, ranked second. Qualitatively, SR-DBM produced the smallest residual errors and best preserved fine structures and lesions. 
    
    \noindent\textbf{\textit{Significance.}} SR-DBM enables accurate and efficient HR MRI reconstruction from rapid LR acquisitions, supporting improved lesion delineation in brain and prostate imaging while reducing scan time and motion artifacts.
\end{abstract}
\newpage 
\section{Introduction}
Magnetic resonance imaging (MRI) is a cornerstone of clinical diagnosis and biomedical research owing to its excellent soft-tissue contrast and ability to generate multiple contrasts without ionizing radiation. Quantitative techniques, such as magnetization-prepared two rapid acquisition gradient echo (MP2RAGE) T1 mapping, yield measurements that are largely free of reception bias and transmit-field inhomogeneity~\cite{marques2010mp2rage}, while T2-weighted (T2w) imaging provides the tissue contrast needed to delineate tumor boundaries in prostate cancer~\cite{hoeks2011prostate}. A common drawback of these acquisitions is their length, which increases patient discomfort and the risk of motion artifacts and forces a trade-off between spatial resolution and scan time~\cite{safari2026systematic}. Increasing the voxel size shortens the acquisition time but introduces partial volume effects that degrade diagnostic quality, motivating super-resolution (SR) as a means of recovering high-resolution (HR) detail from rapidly acquired low-resolution (LR) images.

SR is an ill-posed inverse problem that is classically addressed within a maximum a posteriori framework combining a data-fidelity term with a hand-crafted regularizer such as total variation or wavelet sparsity~\cite{lepcha2023image}. Deep learning has largely superseded these approaches, learning the LR-to-HR mapping directly from data and achieving markedly higher reconstruction quality~\cite{khateri2025mri}. Generative adversarial networks (GANs) produce visually sharp results but are prone to mode collapse and unstable training, which can compromise their reliability in clinical use~\cite{yi2019generative}; structure-preserving variants such as SPSR incorporate gradient guidance to mitigate geometric distortion~\cite{ma2020structure}. More recently, transformer and state-space architectures, including SwinIR~\cite{liang2021swinir} and MambaIR~\cite{guo2024mambair}, have improved long-range dependency modeling for image restoration.

Diffusion models have emerged as a powerful class of generative models and have shown strong performance across MRI reconstruction~\cite{safari2025self}, synthesis~\cite{pan2025cycle, mardhekar2026residual}, and SR~\cite{chang2024high, Safari_2025_res_SRDiff}. Two limitations, however, hinder their direct use for restoration: the iterative reverse process requires many sampling steps and is computationally demanding, and the reliance on a pure Gaussian prior to initialize the reverse process is better suited to unconditional generation than to recovering an image from a degraded measurement. To address the latter limitation, residual-shifting methods such as ResShift~\cite{yue2024efficient} and Res-SRDiff~\cite{Safari_2025_res_SRDiff} center the reverse process on the LR image, and bridge-based methods such as the image-to-image Schr\"odinger bridge (I\textsuperscript{2}SB)~\cite{liu20232} construct a stochastic path that directly connects the LR and HR distributions. These formulations improve efficiency and restoration fidelity, yet they are typically defined over discrete Markov chains or rely on nonlinear bridges that still require comparatively long sampling trajectories.

In this study, we introduce super-resolution diffusion bridge model (SR-DBM), a diffusion bridge model that formulates MRI SR as a continuous-time stochastic transport between the LR and HR image distributions. SR-DBM defines a mean-reverting stochastic differential equation (SDE) whose endpoints are pinned to the paired LR and HR images through Doob's $h$-transform, so that reconstruction is initialized from the measured anatomy rather than from an uninformative Gaussian prior, and the HR image is recovered by integrating the associated reverse-time dynamics with a learned score function. We evaluate SR-DBM on ultra-high-field 7T brain T1 MP2RAGE maps and pelvic T2w prostate images and benchmark it against nine representative methods spanning interpolation, GAN, transformer, state-space, and diffusion-based techniques. The main contributions of this work are as follows:

\begin{itemize}
  \item We instantiate the generalised residual diffusion bridge~\cite{wang2026residualrdbm} for MRI super-resolution as a mean-reverting Ornstein-Uhlenbeck process whose stationary mean is the LR image, pinned to the paired HR image by Doob's $h$-transform, and give the   resulting bridge coefficients in closed form. This makes residual-shifting diffusion~\cite{yue2024efficient,Safari_2025_res_SRDiff} the discrete-time special case of a continuous-time construction, and initialises the reverse process from the LR measurement   rather than from Gaussian noise.
  
  \item We derive a deterministic non-Markovian sampler for this bridge that reconstructs the HR image in ten network evaluations, and we characterise reconstruction quality and inference cost as a function of the number of sampling steps. 
  
  \item We evaluate the method on two anatomies and contrasts, including 7\,T brain T1 MP2RAGE maps and pelvic T2w prostate images, against nine baselines spanning interpolation, GAN, transformer, state-space and diffusion families, with per-patient paired statistics and effect sizes.
\end{itemize}

\section{Materials and Methods}

In this section, we first review the necessary background, progressing from the standard denoising diffusion probabilistic model (DDPM) to diffusion bridges, and finally introduce our proposed super-resolution diffusion bridge model (SR-DBM), which recovers an HR image $\hr$ from its LR counterpart $\lr$. As in Safari et al.~\cite{Safari_2025_res_SRDiff} and Yue et al.~\cite{yue2024efficient}, both images are assumed to share the same spatial grid, obtained by pre-upsampling the LR image to the HR matrix size using nearest-neighbor interpolation. Throughout, we denote the HR-over-LR residual by $\rs = \hr - \lr$.

\subsection{Background: from DDPM to diffusion bridges}
Denoising diffusion probabilistic models (DDPMs) consist of a forward process that gradually corrupts a clean image into Gaussian noise $\mathcal{N}(\bz,\bi)$ over $T$ steps, and a reverse process in which a neural network learns to invert this trajectory. Because the reverse process starts from pure Gaussian noise and typically requires many steps, DDPMs are inefficient for restoration tasks such as super-resolution (SR), where the LR image already provides a strong conditioning signal.

The DDPM has a continuous-time counterpart in which the forward corruption and its time reversal are described by stochastic differential equations (SDEs), with the reverse dynamics driven by the score function $\nabla_{\mathbf{x}_t}\log p(\mathbf{x}_t)$~\cite{song2020score}. Applying a Doob's $h$-transform to such an SDE removes the requirement that the process terminates at a Gaussian prior and instead pins it to a prescribed endpoint, yielding a \emph{diffusion bridge} between two chosen distributions. The recently proposed residual diffusion bridge model (RDBM)~\cite{wang2026residualrdbm} unifies this construction. It combines a mean-reverting Ornstein-Uhlenbeck (OU) process with Doob's $h$-transform and shows that many existing bridges are recovered as special cases. Building on this framework, SR-DBM extends the generalized bridge to MRI SR by pinning a mean-reverting OU process, whose stationary mean is the LR image, to the paired HR image, thereby extending residual-shifting diffusion~\cite{yue2024efficient,Safari_2025_res_SRDiff} from a discrete Markov chain to a continuous-time bridge. For consistency with the DDPM notation above, we present the model in discrete time over $T$ steps, corresponding to a time discretization of the underlying $h$-transformed SDE (\figurename~\ref{fig:flowchart}).

\subsection{Super-resolution diffusion bridge model}

\begin{figure}[t]
	\centering
	\includegraphics[width=\textwidth]{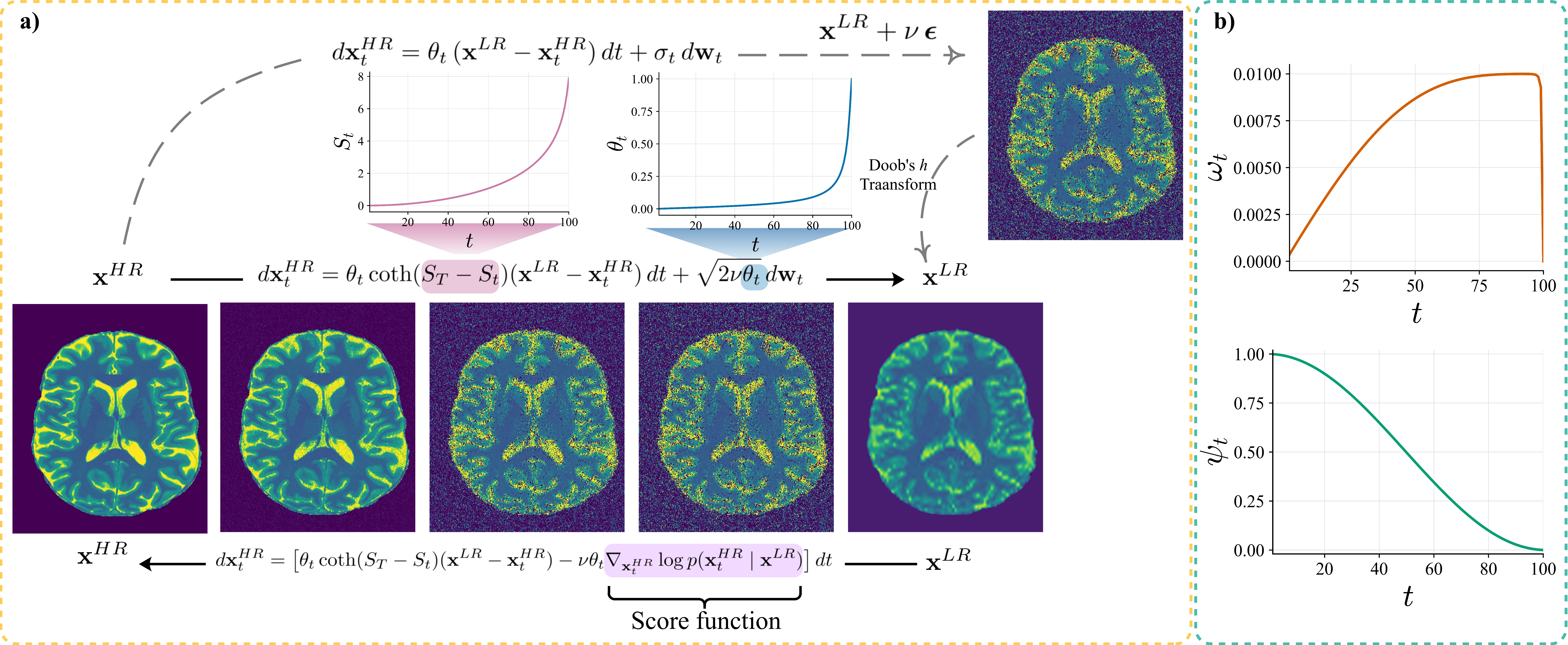}
	\caption{Overview of the proposed SR-DBM. (a) The forward process transports the HR image $\hr$ toward the LR image $\lr$ along a diffusion bridge obtained as a Doob's $h$-transform of a mean-reverting stochastic differential equation whose stationary mean is $\lr$, pinning the trajectory to $\hr$ at $t=0$ and to $\lr$ at $t=T$. The reverse process integrates the corresponding reverse-time dynamics to reconstruct $\hr$ from $\lr$. The image panels illustrate the diffused state at representative time steps, progressing from the HR image on the left to the LR image on the right. (b) The mean coefficient $\psi_t$ decreases monotonically from $1$ to $0$, while the noise scale $\omega_t$ rises and then decays, vanishing at both endpoints, as determined by the drift schedule and the global noise level $\nu$.}
	\label{fig:flowchart}
\end{figure}

Instead of diffusing $\hr$ toward a Gaussian prior, SR-DBM constructs a \emph{diffusion bridge} anchored at the HR image at $t=0$ and at the LR image at $t=T$ (\figurename~\ref{fig:flowchart}). Its backbone is a mean-reverting OU process whose stationary mean is the LR image,
\begin{equation}
	d\mathbf{x}_t = \theta_t\,(\lr-\mathbf{x}_t)\,dt + \sqrt{2\nu\,\theta_t}\;d\mathbf{w}_t,
	\label{eq:ou}
\end{equation}
where $\{\theta_t\}_{t=1}^{T}$ is a positive drift schedule (Section~\ref{sec:scheduler}), $\nu$ is a hyper-parameter controlling the global noise level, and $\mathbf{w}_t$ is a standard Wiener process. Applying a Doob's $h$-transform to \eqref{eq:ou} removes the terminal stationary noise and conditions the process to reach the fixed endpoint $\mathbf{x}_T=\lr$, turning the OU process into a bridge between $\hr$ and $\lr$~\cite{wang2026residualrdbm}. This bridge admits a tractable Gaussian marginal at every time step, so that any intermediate state can be sampled in closed form.

\subsubsection{Forward process}
Let $S_t=\sum_{k=1}^{t}\theta_k$ denote the cumulative drift, with $S_0=0$ and total mass $S_T$; here $S_t$ and $S_T-S_t$ are the discrete counterparts of $\int_0^t\theta_z\,dz$ and $\int_t^T\theta_z\,dz$, respectively. The bridge is characterized by a mean coefficient $\psi_t$ and a noise scale $\omega_t$,
\begin{equation}
	\psi_t \;=\; \frac{\sinh\!\big(S_T-S_t\big)}{\sinh\!\big(S_T\big)},
	\qquad
	\omega_t^{2} \;=\; 2\nu\,
	\frac{\sinh\!\big(S_t\big)\,\sinh\!\big(S_T-S_t\big)}{\sinh\!\big(S_T\big)}.
	\label{eq:coeffs}
\end{equation}
By construction, the mean coefficient decreases monotonically from $\psi_0=1$ to $\psi_T=0$, and the noise scale vanishes at both endpoints, $\omega_0=\omega_T=0$, so that the trajectory is anchored to $\hr$ and $\lr$ without residual noise (\figurename~\ref{fig:flowchart}b). Consequently, the marginal distribution of the diffused state $\mathbf{x}_t$ is
\begin{equation}
	q\big(\mathbf{x}_t \,\big|\, \hr,\lr\big)
	= \mathcal{N}\!\big(\mathbf{x}_t;\;
	\lr + \psi_t\,\rs,\;\omega_t^{2}\bi\big),
	\label{eq:marginal}
\end{equation}
whose mean, equivalently $\psi_t\,\hr + (1-\psi_t)\,\lr$, is a convex interpolation that transports the HR image ($t\!=\!0$) toward the LR image ($t\!=\!T$). Analogously, using the reparameterization trick,
\begin{equation}
	\mathbf{x}_t = \lr + \psi_t\,\rs + \omega_t\,\boldsymbol{\epsilon},
	\qquad \boldsymbol{\epsilon}\sim\mathcal{N}(\bz,\bi).
	\label{eq:reparam}
\end{equation}

\subsubsection{Reverse process and training objective}


The reverse process reconstructs $\hr$ from $\lr$ by learning a network $g_\phi$ that inverts the bridge. Its starting point is fixed by the forward marginal: evaluating \eqref{eq:marginal} at $t=T$ with $\psi_T=0$ and $\omega_T=0$ gives $q(\mathbf{x}_T\mid\hr,\lr)=\mathcal{N}(\mathbf{x}_T;\lr,\bz)=\delta(\mathbf{x}_T-\lr)$. Since the dependence on $\hr$ vanishes, this terminal distribution depends on $\lr$ alone, which we write $p(\mathbf{x}_T\mid\lr)$. The posterior over the clean image then factorizes as

\begin{equation}
	p_\phi\big(\hr\!\mid\!\lr\big)
	=\!\int p\big(\mathbf{x}_T\!\mid\!\lr\big)
	\prod_{t=1}^{T} p_\phi\big(\mathbf{x}_{t-1}\!\mid\!\mathbf{x}_t,\lr\big)
	\,d\mathbf{x}_{1:T},
\end{equation}
with Gaussian reverse kernels $p_\phi(\mathbf{x}_{t-1}\!\mid\!\mathbf{x}_t,\lr)=\mathcal{N}(\boldsymbol{\mu}_\phi,\boldsymbol{\Sigma}_\phi)$ and $p\big(\mathbf{x}_T\!\mid\!\lr\big)=\delta(\mathbf{x}_T-\lr)$. Training minimizes the Kullback-Leibler divergence between each kernel and the tractable forward posterior $q(\mathbf{x}_{t-1}\!\mid\!\mathbf{x}_t,\hr,\lr)$ obtained from \eqref{eq:marginal}. Because this posterior mean is an affine function of the clean image, the divergence reduces, after dropping the time-dependent weights, to predicting the clean HR image $\hr$ from $\mathbf{x}_t$. To ease optimization, we let the network predict a \emph{residual correction} to the LR condition,
\begin{equation}
	\hathr = g_\phi(\mathbf{x}_t,\lr,t) = \lr + f_\phi(\mathbf{x}_t,\lr,t),
	\label{eq:residual_pred}
\end{equation}
so that $f_\phi$ estimates the HR-over-LR residual $\rs$. The overall objective combines an $\ell_1$ data-fidelity term with a learned perceptual image patch similarity (LPIPS) loss $\ell_p$,
\begin{equation}
	\mathcal{L}_\phi = \lambda\,
	\big\lVert g_\phi(\mathbf{x}_t,\lr,t) - \hr\big\rVert_{1}
	+ \ell_p\big(g_\phi(\mathbf{x}_t,\lr,t),\hr\big),
	\label{eq:loss}
\end{equation}
where $\lambda$ balances the two terms and was set to $\lambda=4$ in this study. The complete training procedure is summarized in Algorithm~\ref{alg:training}.

\RyanEdits{
\begin{algorithm}[t]
	\caption{Training}
	\label{alg:training}
	\begin{algorithmic}[1]
		\Require paired data $(\hr,\lr)$; steps $T$; weight $\lambda$; network $g_\phi$
		\Repeat
		\State $(\hr,\lr)\sim p_{\mathrm{data}}$
		\State $t\sim\mathcal{U}\{1,\dots,T\}$
		\State $\boldsymbol{\epsilon}\sim\mathcal{N}(\bz,\bi)$
		\State $\mathbf{x}_t^{HR}\gets \psi_t\,\hr+(1-\psi_t)\,\lr+\omega_t\,\boldsymbol{\epsilon}$
		\Comment{forward marginal, Eq.~\eqref{eq:marginal}}
		\State $\hathr\gets g_\phi\!\left(\mathbf{x}_t^{HR},\lr,t\right)$
		\State $\mathcal{L}\gets \lambda\,\big\lVert \hathr-\hr\big\rVert_1
		+ \ell_{p}\!\left(\hathr,\hr\right)$
		\State take a gradient-descent step on $\nabla_\phi\mathcal{L}$
		\Until{converged}
	\end{algorithmic}
\end{algorithm}
}

\subsubsection{Sampling}
Reconstruction starts from the LR image, $\mathbf{x}_T=\lr$, and follows a deterministic (non-Markovian) reverse trajectory over a strided subset of $S\ll T$ time points for efficiency. Let $t$ denote the current time and $s<t$ the next (smaller) time on the sampling grid. At each step the network predicts the clean HR image $\hathr=g_\phi(\mathbf{x}_t,\lr,t)$, and the state is updated in closed form as
\begin{equation}
	\mathbf{x}_{s}=\lr
	+\frac{\omega_{s}}{\omega_t}\big(\mathbf{x}_t-\lr\big)
	+\Big(\psi_{s}-\psi_t\frac{\omega_{s}}{\omega_t}\Big)
	\big(\hathr-\lr\big).
	\label{eq:sampling}
\end{equation}
Two boundary cases follow. The final step ($s=0$) is obtained by direct substitution in \eqref{eq:sampling}, giving $\mathbf{x}_{0}=\hathr$ (since $\psi_0=1$, $\omega_0=0$), which is the reconstructed HR image. At the first step, the terminal state $\mathbf{x}_T=\lr$ is noise-free ($\omega_T=0$), so no stochastic component is propagated and the update reduces to $\mathbf{x}_{s}=\lr+\psi_{s}(\hathr-\lr)$. The full procedure is summarized in Algorithm~\ref{alg:sampling}.

\RyanEdits{
\begin{algorithm}[t]
	\caption{Sampling}
	\label{alg:sampling}
	\begin{algorithmic}[1]
		\Require LR image $\lr$; trained network $g_\phi$;
		sampling grid $T=t_1>t_2>\dots>t_S\geq 1$, $t_{S+1}=0$
		\State $\mathbf{x}_{t_1}^{HR}\gets \lr$
		\Comment{$\omega_T=0\Rightarrow \mathbf{x}_T^{HR}=\lr$}
		\For{$i=1$ \textbf{to} $S$}
		\State $t\gets t_i,\quad s\gets t_{i+1}$
		\State $\hathr\gets g_\phi\!\left(\mathbf{x}_t^{HR},\lr,t\right)$
		\Comment{predict clean HR}
		\If{$s=0$}
		\State $\mathbf{x}_{s}^{HR}\gets \hathr$
		\ElsIf{$i=1$}
		\State $\mathbf{x}_{s}^{HR}\gets \lr+\psi_s\big(\hathr-\lr\big)$
		\Else
		\State $\mathbf{x}_{s}^{HR}\gets \lr
		+\dfrac{\omega_s}{\omega_t}\big(\mathbf{x}_t^{HR}-\lr\big)
		+\Big(\psi_s-\psi_t\dfrac{\omega_s}{\omega_t}\Big)\big(\hathr-\lr\big)$
		\EndIf
		\EndFor
		\State \Return $\mathbf{x}_0^{HR}$
	\end{algorithmic}
\end{algorithm}
}

\subsection{Noise scheduler}\label{sec:scheduler}
We reuse the cosine schedule of Nichol and Dhariwal~\cite{nichol2021improved} to set the OU drift $\{\theta_t\}_{t=1}^{T}$. Defining $\bar{\alpha}_t=\cos^2\!\big(\tfrac{t/T+0.008}{1.008}\cdot\tfrac{\pi}{2}\big)$, we set $\theta_t=\min\!\big(1-\bar{\alpha}_t/\bar{\alpha}_{t-1},\,0.999\big)$, which is positive and yields a smoothly increasing cumulative drift $S_t$. Together with the global noise level $\nu=1\times10^{-4}$, this fixes the bridge coefficients in \eqref{eq:coeffs}: $\psi_t$ decreases monotonically from $1$ to $0$, while $\omega_t$ rises and then decays to zero at both endpoints (\figurename~\ref{fig:flowchart}b). The peak noise scale admits the closed form $\max_t\omega_t=\sqrt{\nu\tanh(S_T/2)}\le\sqrt{\nu}$;\footnote{Maximising \eqref{eq:coeffs} at $S_t=S_T/2$ and using $\sinh(2x)=2\sinh x\cosh x$ gives $\omega^2=2\nu\sinh^2(S_T/2)/\sinh(S_T)=\nu\tanh(S_T/2)$.} for the present schedule $S_T=7.88$, so $\max_t\omega_t=1.0\times10^{-2}$, attained near $t=92$. The forward process is therefore dominated by the convex interpolation between $\hr$ and $\lr$, and the reverse trajectory used at inference is deterministic. We used $T=100$ diffusion steps for training
and $S=10$ strided steps for sampling.

\subsection{Network architecture}
Following seminal DDPM studies, $g_\phi$ adopts a convolutional U-Net backbone. The LR condition $\lr$ is concatenated channel-wise with the diffused input $\mathbf{x}_t$, and  the diffusion step $t$ is injected through a sinusoidal embedding followed by a multilayer perceptron that modulates each residual block (feature-wise scale/shift). Each residual block uses weight-standardized convolutions with group normalization and SiLU activations. The encoder and decoder comprise four resolution levels with channel multipliers $(1,2,4,8)$ and a base width of $64$. An efficient linear-attention layer is applied at every scale, and a full self-attention layer is placed at the bottleneck. Skip connections link matching encoder/decoder scales, and the final prediction is added to $\lr$ as described above.

\subsection{Implementation details}
The model was implemented in PyTorch and trained on a single NVIDIA A6000 Ada GPU. We used $T=100$ training steps and $S=10$ sampling steps, a batch size of $16$, and the Adam optimizer with a learning rate of $2\times10^{-4}$, $1000$ warm-up steps, and cosine decay. An exponential moving average (decay $0.9995$) of the weights was maintained for inference, and images were normalized to $[-1,1]$. The brain and prostate models were each trained for 100 epochs.

\subsection{Patient data acquisition and preprocessing}

Two datasets spanning distinct anatomies and contrasts were used to train and evaluate the proposed framework: an institutional ultra-high-field 7T brain T1 MP2RAGE cohort~\cite{middlebrooks20247} and the publicly available ProstateX axial 3\,T T2w prostate cancer dataset~\cite{armato2018prostatex}. Together, these data enabled a comprehensive assessment across both neuroimaging and oncologic applications.

The institutional brain cohort comprised 142 patients with confirmed multiple sclerosis. Data were retrospectively collected under Mayo Clinic IRB approval and anonymized in accordance with institutional policies. Patients were partitioned into non-overlapping training (121 patients, 14{,}566 axial slices) and testing (21 patients, 2{,}552 slices) subsets. Imaging was performed using a T1 MP2RAGE protocol; full acquisition parameters are summarized in Table~\ref{tab:imaging_params}. Brain masks were derived from the inversion-1 images using FSL BET~\cite{smith2002fast} and applied to the T1 MP2RAGE maps to suppress extracranial signal. For model input, the T1 maps were down-sampled by a factor of 4 along each spatial dimension, yielding a voxel size of $3.2 \times 3.2 \times 3.2$ mm\textsuperscript{3}.

For the prostate cohort, 334 patients were randomly selected from the ProstateX dataset and split into non-overlapping training (268 patients, 10,480 slices) and evaluation (66 patients, 2,668 slices) sets. T2-weighted images were acquired as summarized in Table~\ref{tab:imaging_params}. The T2w images were down-sampled by factors of 9 in-plane and 2 through-plane, producing an effective voxel size of $2 \times 2 \times 3$ mm\textsuperscript{3}.

\begin{table}[tb!]
\centering
\caption{MRI acquisition parameters for the institutional brain (multiple sclerosis) and ProstateX cohorts.}
\label{tab:imaging_params}
\resizebox{\textwidth}{!}{%
\begin{tabular}{lll}
\toprule
Parameter & Brain (MS) & Prostate (ProstateX) \\
\midrule
Scanner                          & Siemens MAGNETOM Terra (7\,T)                        & Siemens 3\,T (Trio / Skyra) \\
RF coil                          & 8-ch Tx / 32-ch Rx head                              & --- \\
Sequence                         & MP2RAGE (T1 map)                                         & T2w \\
Repetition time (TR)             & 4.5\,s                                               & 5.6\,s \\
Echo time (TE)                   & 2.2\,ms                                              & 104\,ms \\
Inversion times (TI$_1$/TI$_2$)  & 0.95 / 2.5\,s                                        & --- \\
Flip angle(s)                    & 6$^\circ$ / 4$^\circ$                                & 110$^\circ$ \\
Field of view (FOV)              & 230 $\times$ 230\,mm\textsuperscript{2}              & 192 $ \times $ 192 \\
Matrix size                      & 288 $\times$ 288                                     & 256 $\times$ 256 \\
Acquired resolution              & 0.8 $\times$ 0.8 $\times$ 0.8\,mm\textsuperscript{3} & 0.66 $\times$ 0.66 $\times$ 1.5\,mm\textsuperscript{3} \\
Total scan time                  & 8:44\,min                                            & --- \\
\bottomrule
\multicolumn{3}{l}{--- denotes a parameter not reported for that cohort.}
\end{tabular}
}
\end{table}

For both cohorts, data preparation followed the pipeline described in~\cite{safari2026efficient}. All images were intensity-normalized to the range $[-1, 1]$ using the 1st and 99th percentiles, thereby limiting the influence of outliers while preserving tissue contrast. To establish a controlled baseline across methods, no data augmentation was applied during training. Slices dominated by background were excluded prior to training and evaluation: for the brain data, the first and last five slices of each volume were removed together with any slice containing $>95\%$ background, and for the prostate data, slices with predominantly background content were similarly discarded.

\subsection{Quantitative and statistical analysis}

We benchmarked the proposed method against nine comparison approaches: Bicubic interpolation, Pix2pix~\cite{isola2017image}, CycleGAN~\cite{zhu2017unpaired}, SPSR~\cite{ma2020structure}, I\textsuperscript{2}SB~\cite{liu20232}, SwinIR~\cite{liang2021swinir} and MambaIR~\cite{guo2024mambair}, SR-EMamba~\cite{safari2026efficient}, and Res-SRDiff~\cite{Safari_2025_res_SRDiff}.

The quality of the reconstructed HR image $\hathr$ was quantified using four metrics: peak signal-to-noise ratio (PSNR), structural similarity index (SSIM)~\cite{wang2004image}, gradient magnitude similarity deviation (GMSD)~\cite{xue2013gradient}, and learned perceptual image patch similarity (LPIPS)~\cite{zhang2018unreasonable}. PSNR quantifies the voxel-wise residual error on a logarithmic scale (higher is better)~\cite{Safari2023_medfusiongan}; although it is widely reported, it correlates only weakly with perceived image quality, which motivates the complementary metrics used here. SSIM captures local structural agreement (higher is better), GMSD summarises the consistency of gradient magnitude and is therefore sensitive to edge and boundary fidelity (lower is better), and LPIPS measures distance in the feature space of a pretrained network and is the most closely aligned with human perceptual judgements (lower is better). Together, PSNR and SSIM characterise distortion, GMSD characterises gradient structure, and LPIPS characterises learned perceptual similarity.

For each metric-method combination, descriptive statistics were summarized as the mean and standard deviation. Because the metric distributions were not assumed to be Gaussian, comparisons were performed non-parametrically. For each metric, every competing method was compared against the proposed method using a two-sided Wilcoxon signed-rank test, and the resulting $p$-values were adjusted within each metric using the Holm correction for multiple comparisons. All analyses were conducted in \texttt{R} (version 4.3), with statistical significance defined as $p < 0.05$.

\section{Results}
We benchmarked the proposed method against the nine comparison approaches on the 7T brain T1 MP2RAGE maps and the pelvic T2w prostate images. Reconstruction quality was assessed quantitatively and qualitatively through visual inspection of the reconstructions and their difference maps, with statistical significance reported alongside the quantitative comparisons. The findings for the two anatomical regions are presented separately in section~\ref{subsec:brain} for the brain T1 MP2RAGE maps and in section~\ref{subsec:pelvic} for the pelvic T2w prostate images.

\subsection{Brain T1 MP2RAGE maps}\label{subsec:brain}
\begin{figure}[!tbh]
	\centering
	\includegraphics[width=\textwidth]{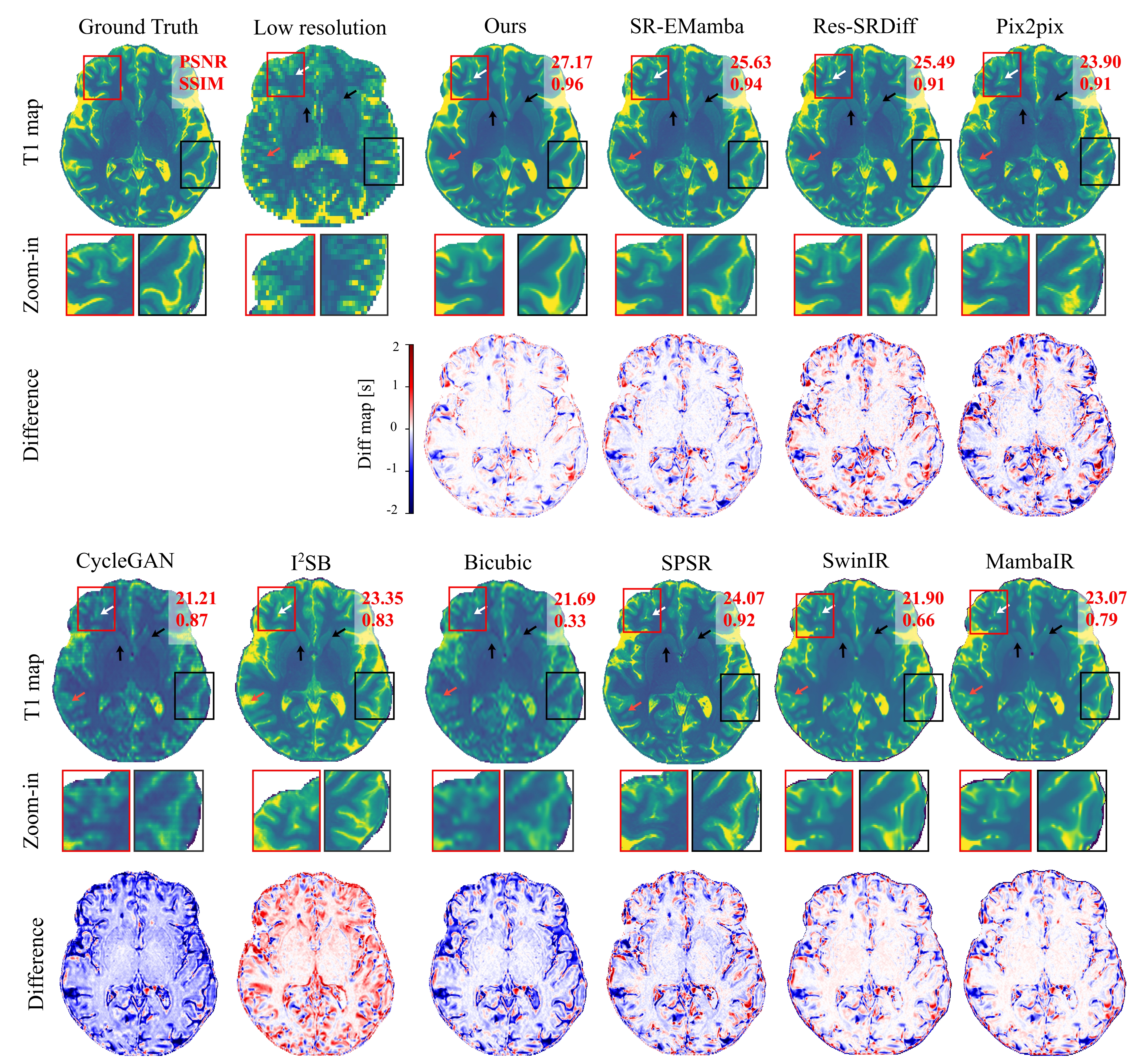}
	\caption{Qualitative results for the 7T brain T1 MP2RAGE maps. The first row shows the ground truth and the low-resolution input together with the reconstructions from the proposed method and the comparative models, with the PSNR and SSIM of each reconstruction reported in the upper-right corner. The zoomed-in regions corresponding to the red and black boxes, where the white, black, and red arrows highlight areas in which the proposed method better recovers fine structures. The difference maps illustrate the spatial deviations of the reconstructions from the ground truth.}
	\label{fig:qual_brain}
\end{figure}

The proposed method reconstructed the 7T brain T1 MP2RAGE maps with higher structural fidelity and lower residual error than the competing approaches, as illustrated in figure~\ref{fig:qual_brain}. The first row presents the ground truth (HR) and the LR starting point alongside the reconstruction produced by each method, the second row shows magnified views of the regions delineated by the red and black boxes, and the third row displays the corresponding difference maps. For the representative slice, the proposed method achieved the highest PSNR and SSIM (27.17 dB and 0.96), and recovered fine cortical and deep gray matter structures more accurately than the other methods, as indicated by the white, black, and red arrows. The difference maps corroborate these findings, with the proposed method yielding the smallest and most spatially uniform residuals, whereas CycleGAN and Bicubic showed a marked global bias, and I\textsuperscript{2}SB produced larger errors throughout the parenchyma.`

\begin{figure}[t]
	\centering
	\includegraphics[width=\linewidth]{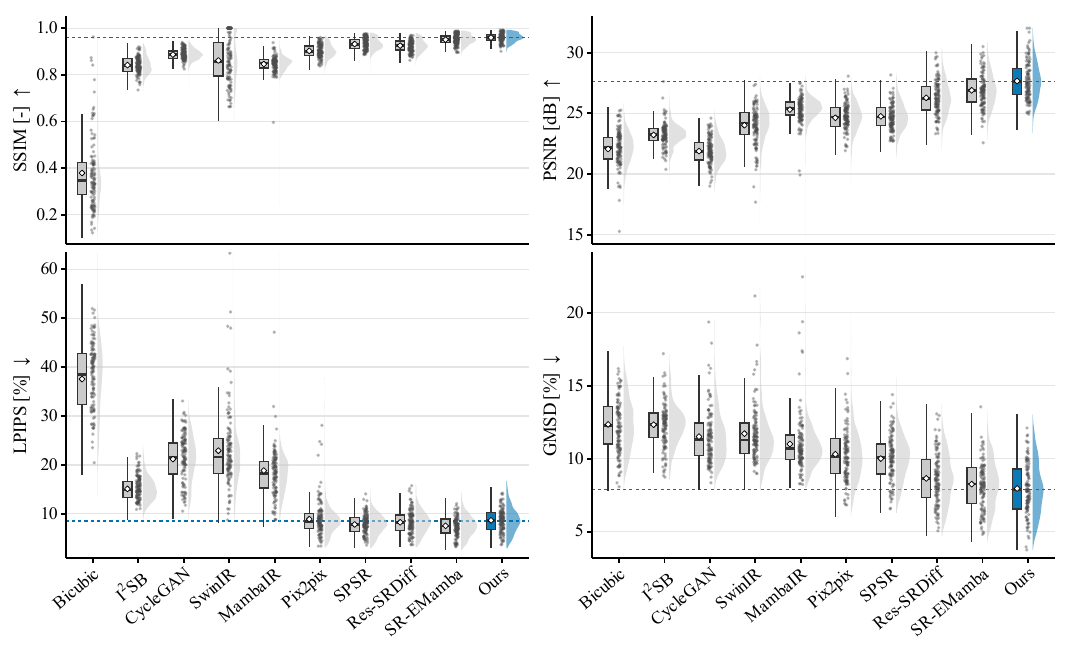}
    \caption{Distribution of the quantitative metrics for the 7\,T brain T1 MP2RAGE maps across all methods. Arrows in the axis labels give the direction of better performance. The dashed horizontal line marks the median of our method to ease comparison.}
	\label{fig:box_brain}
\end{figure}

These observations are consistent with the quantitative metrics reported in Table~\ref{tab:brain}. The proposed method attained the highest PSNR (27.66 $\pm$ 1.52 dB) and SSIM (0.96 $\pm$ 0.02) and the lowest GMSD (7.96 $\pm$ 1.86 \%), followed by SR-EMamba (26.90 $\pm$ 1.41 dB, 0.95 $\pm$ 0.02, and 8.26 $\pm$ 1.96). Using the two-sided Wilcoxon signed-rank test with Holm correction, the gains in PSNR, SSIM, and GMSD were statistically significant relative to all comparison methods ($p < 0.05$). For LPIPS, the proposed method (8.69 $\pm$ 2.62 \%) ranked among the leading approaches but was not the best; SR-EMamba, SPSR, and Res-SRDiff obtained lower values  (7.56 $\pm$ 2.17 \%, 7.84 $\pm$ 2.07 \%, and 8.28 $\pm$ 2.36 \%, respectively), each significantly different from the proposed method, whereas the difference with respect to Pix2pix (8.92 $\pm$ 3.53) was not statistically significant ($p = 0.71$). Figure~\ref{fig:box_brain} shows the distribution of each metric across methods, where the proposed method combines a favorable median with a comparatively narrow spread.

\begin{table}[t]
	\centering
	\caption{Quantitative comparison of super-resolution methods on the 7\,T brain T1 MP2RAGE maps. Results are reported as the mean with the standard deviation given as a subscript. Bold and underlined values indicate the best and second-best performance, respectively, and the arrows denote the direction of better results.}
	\label{tab:brain}
    \resizebox{0.7\textwidth}{!}{%
	\begin{tabular}{lcccc}
		\toprule
		Method & PSNR [dB] $\uparrow$ & SSIM [-] $\uparrow$ & GMSD [\%] $\downarrow$ & LPIPS [\%] $\downarrow$ \\
		\midrule
		Bicubic    & $22.03_{\pm 1.36}$ & $0.36_{\pm 0.16}$ & $12.37_{\pm 1.90}$ & $35.55_{\pm 7.66}$ \\
		Res-SRDiff & $26.28_{\pm 1.41}$ & $0.92_{\pm 0.03}$ & $8.65_{\pm 1.71}$ & $8.28_{\pm 2.36}$ \\
		CycleGAN   & $21.89_{\pm 1.09}$ & $0.89_{\pm 0.02}$ & $11.51_{\pm 1.74}$ & $21.19_{\pm 4.62}$ \\
		Pix2pix    & $24.63_{\pm 1.32}$ & $0.90_{\pm 0.03}$ & $10.31_{\pm 1.91}$ & $8.92_{\pm 3.53}$\textsuperscript{a} \\
		SPSR       & $24.76_{\pm 1.12}$ & $0.93_{\pm 0.02}$ & $10.01_{\pm 1.49}$ & $\underline{7.84}_{\pm 2.07}$ \\
		I\textsuperscript{2}SB    & $23.22_{\pm 0.98}$ & $0.84_{\pm 0.04}$ & $12.33_{\pm 1.44}$ & $15.10_{\pm 2.51}$ \\
		SwinIR     & $24.02_{\pm 1.62}$ & $0.86_{\pm 0.10}$ & $11.72_{\pm 2.23}$ & $22.92_{\pm 7.95}$ \\
		MambaIR    & $25.31_{\pm 1.09}$ & $0.85_{\pm 0.04}$ & $11.01_{\pm 1.92}$ & $18.85_{\pm 7.17}$ \\
		SR-EMamba  & $\underline{26.90}_{\pm 1.41}$ & $\underline{0.95}_{\pm 0.02}$ & $\underline{8.26}_{\pm 1.67}$ & $\mathbf{7.56}_{\pm 2.17}$ \\
		Ours       & $\mathbf{27.66}_{\pm 1.52}$ & $\mathbf{0.96}_{\pm 0.02}$ & $\mathbf{7.96}_{\pm 1.86}$ & $8.69{_\pm 2.62}$ \\
		\bottomrule
		\multicolumn{5}{p{\dimexpr0.8\linewidth-2\tabcolsep\relax}}{\footnotesize \textsuperscript{a} Difference relative to the proposed method is not statistically significant (two-sided Wilcoxon signed-rank test with Holm correction, $p > 0.05$); all other differences with respect to the proposed method are significant ($p < 0.05$).}
	\end{tabular}
	}
\end{table}

\subsection{Pelvic T2w images}\label{subsec:pelvic}
\begin{figure}[!tb]
	\centering
	\includegraphics[width=\textwidth]{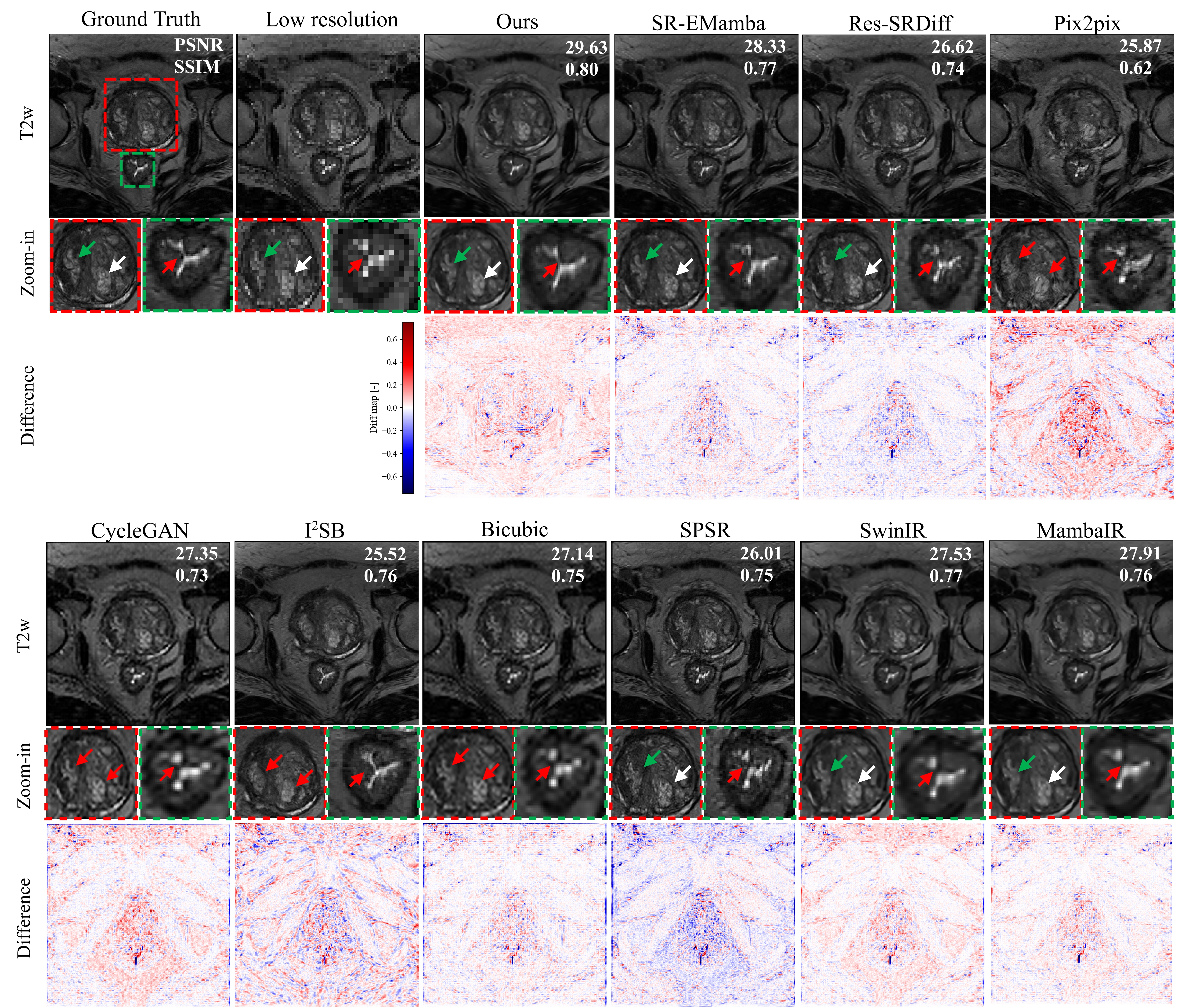}
	\caption{Qualitative results for the pelvic T2w prostate images. The first row shows the ground truth and the low-resolution input together with the reconstructions from the proposed method and the comparative models, with the PSNR and SSIM of each reconstruction reported in the upper-right corner. The zoomed-in regions outlined by the red and green boxes, where the arrows indicate structures and lesions that the proposed method restores closer to the ground truth. The difference maps illustrate the spatial deviations of the reconstructions from the ground truth.}
	\label{fig:qual_pelvis}
\end{figure}

For the pelvic T2w prostate images, the proposed method restored anatomical structures and lesions with improved fidelity to the ground truth, as shown in figure~\ref{fig:qual_pelvis}. The first row displays the ground truth and the low-resolution input together with the reconstructions from each method, the second row provides magnified views of the regions outlined by the red and green boxes, and the third row presents the difference maps. On the representative slice, the proposed method achieved the highest PSNR and SSIM (29.63 dB and 0.80) and reconstructed the lesion and periprostatic tissue closer to the ground truth than the competing methods, as highlighted by the green, white, and red arrows. The difference maps confirm that the proposed method produced the smallest residual error, whereas Pix2pix exhibited a pronounced central bias and Bicubic together with SwinIR retained coarser structural detail.

\begin{figure}[t]
	\centering
	\includegraphics[width=\linewidth]{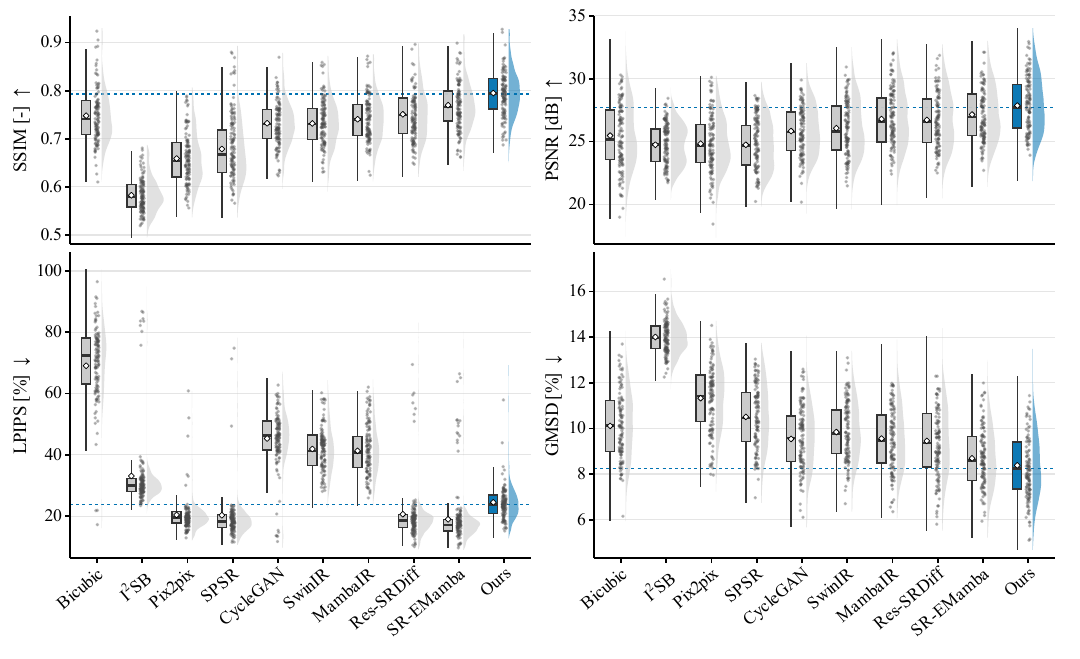}
    \caption{Distribution of the quantitative metrics for the pelvic T2w prostate images across all methods. Arrows in the axis labels give the direction of better performance. The dashed horizontal line marks the median of our method to ease comparison.}

	\label{fig:box_prostate}
\end{figure}

The quantitative results in Table~\ref{tab:pelvic} reflect the same trend. The proposed method obtained the highest PSNR (27.87 $\pm$ 2.29 dB) and SSIM (0.80 $\pm$ 0.05) and the lowest GMSD (8.38 $\pm$ 1.44), ahead of SR-EMamba (27.15 $\pm$ 2.19 dB, 0.77 $\pm$ 0.05, and 8.69 $\pm$ 1.34). Using the two-sided Wilcoxon signed-rank test with Holm correction, these improvements were statistically significant against every comparison method ($p < 0.05$). As with the brain data, the proposed method did not rank first for LPIPS (24.50 $\pm$ 6.53 \%); SR-EMamba (18.96 $\pm$ 9.47), SPSR (20.27 $\pm$ 9.19), Pix2pix (20.38 $\pm$ 5.39), and Res-SRDiff (20.62 $\pm$ 10.90) achieved lower values, and each difference relative to the proposed method was statistically significant ($p < 0.05$). Figure~\ref{fig:box_prostate} presents the distribution of each metric across methods, where the proposed method combines a favorable median with a comparatively narrow spread.

\begin{table}[t]
	\centering
	\caption{Quantitative comparison of super-resolution methods on the pelvic T2w prostate images. Results are reported as the mean with the standard deviation given as a subscript. Bold and underlined values indicate the best and second-best performance, respectively, and the arrows denote the direction of better results.}
	\label{tab:pelvic}
    \resizebox{0.7\textwidth}{!}{%
	\begin{tabular}{lcccc}
		\toprule
		Method & PSNR [dB] $\uparrow$ & SSIM [-] $\uparrow$ & GMSD [\%] $\downarrow$ & LPIPS [\%] $\downarrow$ \\
		\midrule
		Bicubic    & $25.44_{\pm 2.61}$ & $74.52_{\pm 5.64}$ & $10.11_{\pm 1.51}$ & $65.99_{\pm 15.60}$ \\
		Res-SRDiff & $26.72_{\pm 2.26}$ & $75.09_{\pm 5.31}$ & $9.45_{\pm 1.54}$ & $20.62_{\pm 10.90}$ \\
		CycleGAN   & $25.84_{\pm 1.96}$ & $73.23_{\pm 4.64}$ & $9.54_{\pm 1.39}$ & $45.34_{\pm 9.80}$ \\
		Pix2pix    & $24.83_{\pm 2.09}$ & $65.89_{\pm 5.36}$ & $11.33_{\pm 1.38}$ & $20.38_{\pm 5.39}$ \\
		SPSR       & $24.74_{\pm 1.96}$ & $67.87_{\pm 6.68}$ & $10.50_{\pm 1.41}$ & $\underline{20.27}_{\pm 9.19}$ \\
		I\textsuperscript{2}SB    & $24.74_{\pm 1.61}$ & $58.25_{\pm 3.64}$ & $14.00_{\pm 0.76}$ & $33.09_{\pm 12.83}$ \\
		SwinIR     & $26.06_{\pm 2.25}$ & $73.19_{\pm 4.82}$ & $9.84_{\pm 1.31}$ & $41.90_{\pm 7.69}$ \\
		MambaIR    & $26.79_{\pm 2.35}$ & $74.06_{\pm 4.80}$ & $9.55_{\pm 1.40}$ & $41.30_{\pm 7.69}$ \\
		SR-EMamba  & $\underline{27.15}_{\pm 2.19}$ & $\underline{76.98}_{\pm 4.93}$ & $\underline{8.69}_{\pm 1.34}$ & $\mathbf{18.96}_{\pm 9.47}$ \\
		Ours       & $\mathbf{27.87}_{\pm 2.29}$ & $\mathbf{79.51}_{\pm 4.74}$ & $\mathbf{8.38}_{\pm 1.44}$ & $24.50_{\pm 6.53}$ \\
		\bottomrule
        \multicolumn{5}{p{\dimexpr0.8\linewidth-2\tabcolsep\relax}}{\footnotesize All differences with respect to the proposed method are statistically significant (two-sided Wilcoxon signed-rank test with Holm correction, $p<0.05$).}
	\end{tabular}
    }
\end{table}

\section{Discussion}
Prolonged MRI acquisition remains a practical bottleneck in both clinical and research settings, forcing a trade-off between spatial resolution and scan time.  This trade-off motivates super-resolution as a technique to recover high-resolution detail from rapidly acquired low-resolution data. In this work we introduced SR-DBM, a diffusion bridge model that formulates super-resolution as a stochastic transport between the low-resolution and high-resolution image distributions. Rather than initializing the reverse process from pure Gaussian noise, SR-DBM defines a mean-reverting stochastic differential equation whose endpoints are pinned to the paired LR and HR images through Doob's $h$-transform, and it recovers the HR image by integrating the corresponding reverse-time dynamics with a learned score function. Across both the 7T brain T1 MP2RAGE maps and the pelvic T2w prostate images, SR-DBM achieved the highest PSNR and SSIM and the lowest GMSD, and these improvements were statistically significant against every comparison method (Tables~\ref{tab:brain} and~\ref{tab:pelvic}).

The consistent gains in fidelity and structural agreement can be attributed in part to the bridge formulation. Because the generative trajectory is conditioned on the observed LR image at both ends, the reverse process begins from an initialization that already reflects the measured anatomy rather than from an uninformative Gaussian prior, which limits the accumulation of reconstruction noise and yields a more direct mapping from the degraded input to the target. This behavior is visible in the difference maps of figures~\ref{fig:qual_brain} and~\ref{fig:qual_pelvis}, where SR-DBM produced the smallest and most spatially uniform residuals, in contrast to the pronounced global bias of Bicubic and CycleGAN on the brain maps, and the central bias of Pix2pix on the prostate images. Among the diffusion-based baselines, I\textsuperscript{2}SB~\cite{liu20232} is likewise built on a bridge between the LR and HR distributions, yet SR-DBM outperformed it across all fidelity and structural metrics; a plausible explanation is that the mean-reverting drift together with Doob's $h$-transform provides a better conditioned path to the fixed HR endpoint, reducing the noise that can accumulate over the longer sampling trajectories required by the Schr\"odinger bridge. Relative to Res-SRDiff~\cite{Safari_2025_res_SRDiff}, which shifts the residual error over a discrete Markov chain, the continuous-time bridge formulation of SR-DBM offers finer control over the transport and translated into consistent improvements in PSNR, SSIM, and GMSD on both datasets.

The strongest competing method was SR-EMamba, which ranked second on PSNR, SSIM, and GMSD for both anatomies. Nevertheless, SR-DBM retained a statistically significant advantage over SR-EMamba on these fidelity and structural metrics, indicating that the probabilistic bridge recovers voxel intensities and gradient structure more accurately than the deterministic state-space model. The one metric on which SR-DBM did not rank first was LPIPS. SR-EMamba obtained the lowest values on T2w datasets and several other methods including SPSR and Res-SRDiff also achieved lower LPIPS than SR-DBM. On the brain maps, the only comparison that did not reach statistical significance was that between SR-DBM and Pix2pix, whereas on the prostate images every difference relative to SR-DBM was significant. We interpret this pattern in light of the well-documented perception-distortion trade-off~\cite{blau2018perception}, whereby methods that minimize distortion metrics such as PSNR and SSIM tend to incur a penalty on learned perceptual similarity and conversely. SR-DBM is optimized toward faithful reconstruction of the underlying signal, which is reflected in its distortion and gradient-based scores, whereas approaches that emphasize perceptual realism can attain lower LPIPS at the cost of quantitative fidelity. For quantitative T1 mapping and for lesion delineation, where the accuracy of the recovered intensities and boundaries is of primary importance, we regard the fidelity and structural metrics as the more clinically pertinent criteria, although perceptual quality remains relevant for radiological reading and constitutes a target for further improvement.

The improvements afforded by SR-DBM carry practical implications for both applications examined here. For the 7T brain T1 MP2RAGE maps acquired in patients with multiple sclerosis, accurate recovery of tissue intensities and sharp gray-white matter boundaries is essential for the detection and characterization of demyelinating lesions. Furthermore, the reduced GMSD achieved by SR-DBM indicates improved preservation of edge and gradient information. For the pelvic T2w images, the more faithful restoration of the lesion and periprostatic structures, together with the smallest residual error among the evaluated methods, supports more reliable tumor delineation. In both settings, recovering high-resolution detail from faster low-resolution acquisitions may help decrease scan time, thereby reducing the associated risk of patient discomfort or motion artifacts while simultaneously preserving the information required for diagnosis and treatment planning.

Several limitations warrant consideration. Although the datasets are volumetric, SR-DBM was implemented as a two-dimensional method that reconstructs each slice independently, which simplifies training and inference but provides no explicit mechanism to enforce inter-slice continuity; extending the bridge formulation to three dimensions or adding continuity constraints is a natural direction for ensuring volumetric coherence. The perceptual gap reflected in the LPIPS scores suggests a second avenue, namely the incorporation of perceptual or adversarial objectives during training, which could improve perceptual similarity while preserving the fidelity advantages of the bridge. A more detailed characterization of computational cost as a function of the number of sampling steps, and its trade-off against reconstruction quality, would further clarify the efficiency of SR-DBM relative to the diffusion-based baselines. Finally, as with other super-resolution methods, the reconstruction may occasionally alter fine anatomical structure. Integrating uncertainty estimation or structure-preserving mechanisms could help safeguard clinically relevant features and increase confidence in the reconstructed images.

\section{Conclusion}
We introduced SR-DBM, a diffusion bridge model that casts MRI super-resolution as a stochastic transport between the low-resolution and high-resolution distributions, pinned to the paired images through Doob's $h$-transform. On 7T brain T1 MP2RAGE maps and pelvic T2w prostate images, SR-DBM achieved the highest PSNR and SSIM and the lowest GMSD, with statistically significant gains over every comparison method. Future work will extend the formulation to three dimensions and incorporate perceptual objectives to further improve high-resolution MRI reconstruction.

\section*{Conflicts of interest}
There are no conflicts of interest declared by the authors.
\section*{Acknowledgment}
This research is supported in part by the National Institutes of Health under Award Numbers R01DE033512 and R01CA272991.

\section*{Data availability}

The ProstateX dataset is openly accessible through the TCIA portal (\url{https://www.cancerimagingarchive.net/analysis-result/prostatex-seg-hires/}). The institutional dataset used in this study contains sensitive patient information and therefore cannot be released publicly at the time of publication.

\bibliographystyle{unsrt}
\bibliography{./sr_diffusion_bridge.bib}      

\end{document}